\documentclass[pdflatex,sn-mathphys-num]{sn-jnl}

\usepackage{graphicx}%
\usepackage{multirow}%
\usepackage{amsmath,amssymb,amsfonts}%
\usepackage{amsthm}%
\usepackage{mathrsfs}%
\usepackage[title]{appendix}%
\usepackage{xcolor}%
\usepackage{textcomp}%
\usepackage{manyfoot}%
\usepackage{booktabs}%
\usepackage{algorithm}%
\usepackage{algorithmicx}%
\usepackage{algpseudocode}%
\usepackage{listings}%
\usepackage{multirow}
\usepackage{booktabs}
\usepackage{adjustbox}
\usepackage{subfigure}
\usepackage{array}
\usepackage{colortbl}
\newcolumntype{P}[1]{>{\centering\arraybackslash}p{#1}}

\theoremstyle{thmstyleone}%
\theoremstyle{thmstyletwo}%

\theoremstyle{thmstylethree}%

\begin{document}

\title[Article Title]{Symmetry Breaking in Neural Network Optimization: Insights from Input Dimension Expansion}


\author[1]{\fnm{Jun-Jie} \sur{Zhang}}\email{zjacob@mail.ustc.edu.cn}
\equalcont{Both authors contributed equally to this work.}

\author[2,3]{\fnm{Nan} \sur{Cheng}}\email{dr.nan.cheng@ieee.org}

\equalcont{Both authors contributed equally to this work.}

\author[4]{\fnm{Fu-Peng} \sur{Li}}\email{fupengli29@mails.ccnu.edu.cn}

\author[2,3]{\fnm{Xiu-Cheng} \sur{Wang}}\email{xcwang\_1@stu.xidian.edu.cn}

\author[1]{\fnm{Jian-Nan} \sur{Chen}}\email{chennn1994@alumni.sjtu.edu.cn}

\author*[4]{\fnm{Long-Gang} \sur{Pang}}\email{lgpang@ccnu.edu.cn}

\author*[5]{\fnm{Deyu} \sur{Meng}}\email{dymeng@mail.xjtu.edu.cn}

\affil[1]{\orgname{Northwest Institute of Nuclear Technology}, \orgaddress{\street{No. 28 Pingyu Road}, \city{Xi'an}, \postcode{710024}, \state{Shaanxi}, \country{China}}}

\affil[2]{\orgdiv{School of Telecommunications Engineering}, \orgname{Xidian University}, \orgaddress{\street{No. 2 South Taibai Road}, \city{Xi'an}, \postcode{710071}, \state{Shaanxi}, \country{China}}}
\affil[3]{\orgname{State Key Laboratory of ISN}, \orgaddress{\street{No. 2 South Taibai Road}, \city{Xi'an}, \postcode{710071}, \state{Shaanxi}, \country{China}}}

\affil[4]{\orgdiv{Key Laboratory of Quark and Lepton Physics (MOE) \& Institute of Particle Physics}, \orgname{Central China Normal University}, \orgaddress{\street{No. 152 Luoyu Road}, \city{Wuhan}, \postcode{30079}, \state{Hubei}, \country{China}}}

\affil[5]{\orgdiv{School of Mathematics and Statistics and Ministry of Education Key Lab of Intelligent Networks and Network Security}, \orgname{Xi’an Jiaotong University}, \orgaddress{\street{No. 28 Xianning West Road}, \city{Xi'an}, \postcode{710049}, \state{Shaanxi}, \country{China}}}

\abstract{Understanding the mechanisms behind neural network optimization is crucial for improving network design and performance. While various optimization techniques have been developed, a comprehensive understanding of the underlying principles that govern these techniques remains elusive. Specifically, the role of symmetry breaking, a fundamental concept in physics, has not been fully explored in neural network optimization. This gap in knowledge limits our ability to design networks that are both efficient and effective. Here, we propose the symmetry breaking hypothesis to elucidate the significance of symmetry breaking in enhancing neural network optimization. We demonstrate that a simple input expansion can significantly improve network performance across various tasks, and we show that this improvement can be attributed to the underlying symmetry breaking mechanism. We further develop a metric to quantify the degree of symmetry breaking in neural networks, providing a practical approach to evaluate and guide network design.
Our findings confirm that symmetry breaking is a fundamental principle that underpins various optimization techniques, including dropout, batch normalization, and equivariance. By quantifying the degree of symmetry breaking, our work offers a practical technique for performance enhancement and a metric to guide network design without the need for complete datasets and extensive training processes.}




\maketitle
\section{Introduction}

Artificial intelligence (AI) has seen remarkable advancements over the past decade, particularly in neural networks, which have revolutionized fields such as computer vision, natural language processing, and autonomous systems \cite{krizhevsky2012imagenet, 7780459, 10.5555/3295222.3295349, 10.5555/3495724.3495883, mnih2015human}. These models have also made significant strides in scientific computing and creative domains, enhancing areas like healthcare diagnostics, weather forecasting, and art generation \cite{esteva2017dermatologist, rasp2020weatherbench, jumper2021highly, carleo2017solving, 72933a14975149d3a457c57551785b09}.

Despite these successes, the underlying mechanisms of these complex models remain poorly understood \cite{lipton2018mythos}. This lack of interpretability often forces researchers to rely on trial-and-error methods to improve model performance \cite{doshi2017towards}. As neural networks grow in size and complexity, they become increasingly difficult to analyze and understand. The high dimensionality and intricate architectures of these networks pose significant challenges for theoretical analysis. With the continuous growth in data and network parameters, achieving a deeper understanding of these models becomes both more difficult and more crucial \cite{lecun2015deep}.

Given the complexity and expense of directly modeling and analyzing neural networks, it is worth considering more efficient approaches inspired by first principles in scientific fields. One effective strategy is principle-based modeling, which leverages foundational principles from various disciplines to guide the development and optimization of neural networks \cite{doi:10.1073/pnas.2403580121}.

A particularly promising principle is symmetry breaking—a concept deeply rooted in physics that describes how a system that is initially symmetric becomes asymmetric under certain conditions, leading to new and often more stable configurations \cite{onsager1944crystal, anderson1972more}. In the context of neural networks, symmetry breaking \cite{fok2017spontaneous, tanaka2021noether} can be thought of as a mechanism that helps the model escape from local minima and saddle points in the loss landscape \cite{choromanska2015loss, doi:10.1073/pnas.1908636117}, thereby facilitating better optimization and generalization.

In this study, we propose treating symmetry breaking as a fundamental principle and describe its application in neural network optimization. Our investigation began with an intriguing empirical observation: expanding the dimensionality of input pixels and filling the expanded dimensions with constant values significantly improves model performance across various datasets and tasks, including image classification, Physics-Informed Neural Networks (PINNs) \cite{pinn2016}, image coloring, and sentiment analysis.

To understand this phenomenon, we propose the hypothesis of symmetry breaking and claim it to be a fundamental principle in neural network optimization. Our findings indicate that augmenting the input space with additional dimensions of constant values mimics a symmetry breaking process. This augmentation proves advantageous for network training, as it facilitates smoother transitions in parameter values during gradient descent.

To further test and validate the symmetry breaking hypothesis, we conducted a series of experiments comparing the loss landscapes of models trained with different techniques, including equivariance \cite{cohen2016equivariant}, dropout \cite{srivastava2014dropout}, and batch normalization \cite{ioffe2015batchnormalization}. Our findings suggest that these techniques all adhere to the symmetry breaking principle in neural network optimization. Moreover, we discovered that embedding equivariance into the network design is a more effective approach than merely scaling up the network, highlighting a trend that deviates from the current focus on scaling laws.

Finally, inspired by the ``Parasi" scheme, which is used as a metric to measure the ``replica symmetry" of spin-glass systems \cite{parasi1988, liao2024losslandscapeslens} in physics, we developed a procedure to measure the ``degree of symmetry breaking" for the neural network without the need for complete datasets and extensive training processes. This measurement could serve as a metric to evaluate neural network designs.

Our work advances the theoretical understanding of neural network optimization and provides practical tools and methodologies that can be widely applied to improve AI systems across various domains.

\section{Observation: Input Expansion Brings Performance Enhancement}\label{sec2}
\begin{figure}
    \centering
    \includegraphics[width=1\linewidth]{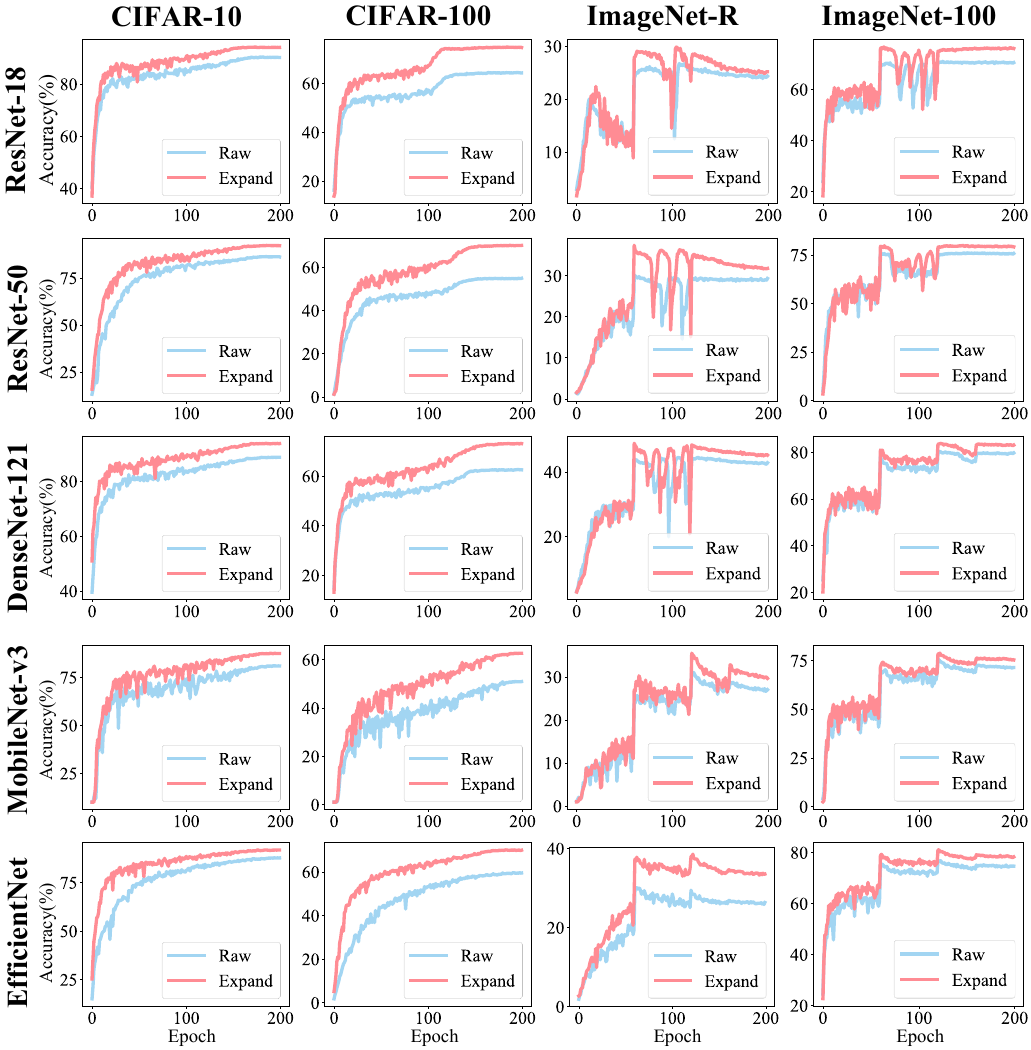}
    \caption{Performance comparison for raw data and expanded data on CIFAR and ImageNet datasets with various models. Each row represents a different dataset: CIFAR-10, CIFAR-100, ImageNet-R, and ImageNet-100, respectively. Each column corresponds to a different model architecture: ResNet-18, ResNet-50, DenseNet-121, MobileNet-v3, and EfficientNet. Within each cell, the plot shows two curves representing test accuracy over epochs: the red curve for expanded data and the blue curve for raw data.}
    \label{fig-cifar}
\end{figure}

\subsection{Impact of Dimension Expansion on Image Classification}

We begin by analyzing image classification performance across various datasets, focusing on models primarily based on Convolutional Neural Network (CNN) architectures. Specifically, we compared the performance of models when fed with original data versus data that had undergone dimension expansion. The datasets used in our experiments included CIFAR-10, CIFAR-100\footnote{https://www.cs.toronto.edu/~kriz/cifar.html}, and ImageNet-R, ImageNet-100\footnote{https://www.image-net.org/}. To ensure a comprehensive evaluation, we employed a range of network architectures, including ResNet-18 and ResNet-50 \cite{7780459} from the classic residual networks, DenseNet-121 \cite{huang2017densely} for deeper network analysis, and EfficientNet \cite{tan2019efficientnet} and MobileNet-v3 \cite{howard2019searching} to assess the impact on lightweight quantized networks. All network architectures adhered to the standard structures provided by Torchvision in PyTorch without any modifications. The dimension expansion technique involved enlarging each dimension of the image by a factor of two. In the expanded image, original pixels were placed at evenly spaced intervals, with additional pixels set to a constant value of 0.5. This method ensured that the original image's information was preserved while introducing additional spatial context. 

\begin{figure}
    \centering
    \includegraphics[width=1\linewidth]{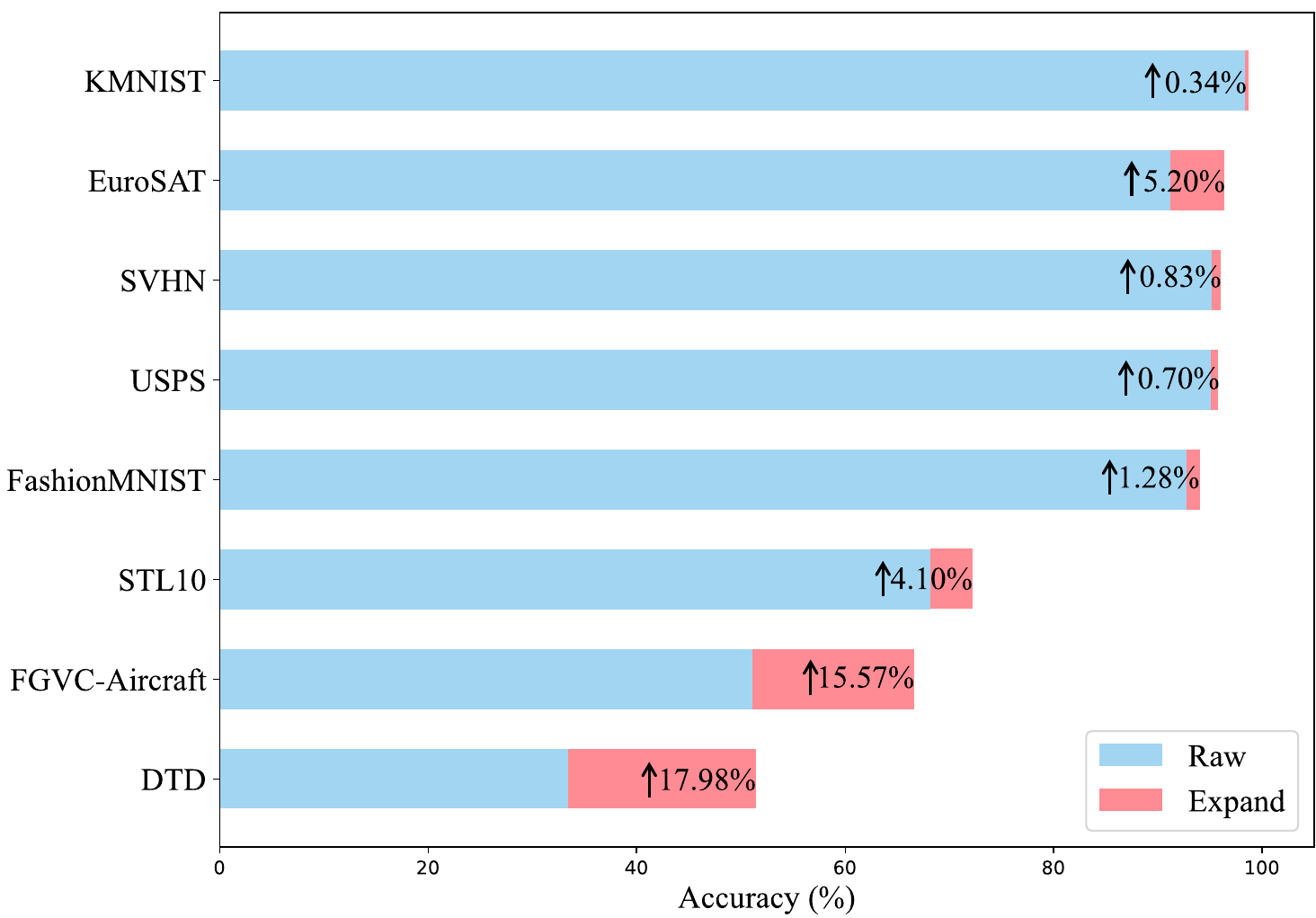}
    \caption{Performance comparison for raw data and expanded data on various datasets using ResNet-18. Each bar represents the test accuracy, with blue bars indicating the accuracy for raw data and red bars showing the additional accuracy gained from using expanded data. The total length of each bar (blue plus red) represents the final test accuracy after applying dimension expansion. Up-arrows next to each dataset indicate the accuracy increment achieved through dimension expansion, with the numerical values beside the arrows quantifying the improvement.}
    \label{fig-more-dataset}
\end{figure}

Our findings, depicted in Fig.~\ref{fig-cifar}, reveal that all models exhibited improved convergence performance in image classification when input data dimensions were expanded. This improvement was reflected in faster training convergence speeds across models, demonstrating the broad applicability of dimension expansion for enhancing model performance. Notably, after data expansion, the classification accuracy of ResNet-18 on CIFAR-10 soared to 94\%, rivaling that of the larger ResNet-101 \cite{7780459}. This result underscores the significant performance gains achieved through image dimension expansion, highlighting its potential to enhance CNN performance across various architectures and datasets.

To further investigate the relationship between data expansion and the intrinsic features of the image datasets, we extended our analysis to include additional datasets covering a broader spectrum: including grayscale datasets such as KMNIST \cite{kmnist} and FashionMNIST \cite{fashionmnist}, thus avoiding any bias introduced by using solely three-channel color images like CIFAR and ImageNet. Given the extensive use of image classification in satellite remote sensing, we incorporated the EuroSAT dataset \cite{eurosat}. Additionally, we selected FGVC-Aircraft \cite{fgvc-aircraft} for its focus on classifying specific aircraft models, which involves distinguishing between highly similar classes, thereby testing the model's feature extraction capabilities. Lastly, the DTD dataset \cite{dtd}, which categorizes textures, was chosen to further validate model performance. This variety ensures a comprehensive evaluation of the effectiveness of the data expansion technique across different types of image data.

Our results, as shown in Fig.~\ref{fig-more-dataset}, demonstrated that data expansion consistently enhances classification performance across all tested datasets. Even for datasets where the baseline accuracy was already high, such as KMNIST, data expansion still led to noticeable improvements in accuracy. For datasets with higher classification difficulty, like FGVC-Aircraft and DTD, the improvements were even more pronounced. Particularly noteworthy is the performance boost observed for FGVC-Aircraft and DTD, which are inherently challenging due to their high intra-class similarity and textural complexity, respectively.

To further validate the effect of data expansion on image classification performance, we evaluated the impact of different expansion factors, ranging from 2 to 5, and various filling strategies, including fixed values (0.0, 0.25, 0.5, 1.0) and random numbers. The experimental results demonstrate that, regardless of the expansion method employed, there is a consistent improvement in image classification performance. Detailed results are provided in the supplementary materials.

\subsection{Impact of Dimension Expansion on AI4Science}

To demonstrate the effects of dimension expansion in the intersection field of AI and scientific discovery, known as ``AI4Science" and ``Science4AI", we performed examples based on PINNs --- a hybrid approach that combines data-driven and physics-driven methods.

\textbf{Data-Driven task --- Equation of State in Quantum Chromodynamics.} Quantum Chromodynamics (QCD) \cite{Gross:1973id, Gross:2022hyw} is the fundamental theory that describes the strong interaction between quarks and gluons. One of the key challenges in QCD is to understand the equation of state (EoS) of the hot and dense QCD matter, which is crucial for explaining phenomena such as the early universe  and the heavy-ion collisions \cite{Yagi:2005yb,Busza:2018rrf,Pang:2016vdc}. Presently, for the hot and dense quark-gluon matter, the QCD EoS can only be obtained by Lattice QCD \cite{Wilson:1974sk,HotQCD:2014kol} calculations based on first principles, which requires huge computational effort. 

Here, we developed a quasi-particle model to reproduce the QCD EoS at zero baryon chemical potential. Our approach utilizes three neural networks to represent the temperature-dependent masses of quasi-quarks and quasi-gluons. 

The neural networks take temperature $T$ as input and masses as outputs. Given these masses, the physical quantities of pressure and energy density can be obtained. Limited by the huge computational cost, we only have 50 points of pressure and energy density from Lattice QCD, hence the task is a date-driven task with 50 Lattice QCD points.

\begin{figure}[h]
    \centering
    \includegraphics[width=1\textwidth]{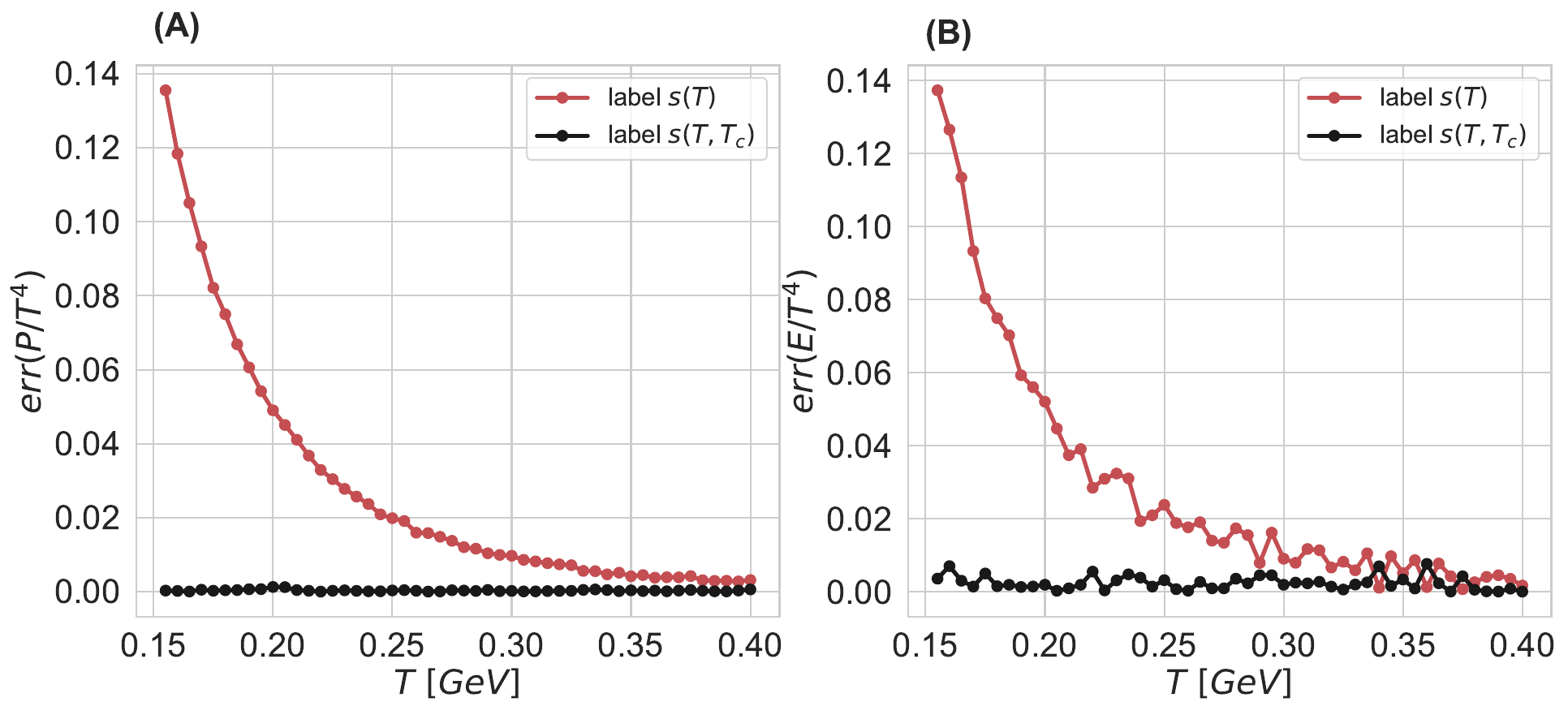}
    \caption{Comparison of the predicted QCD state equation results with expanded (black scatter) and unexpanded (red scatter) input dimensions. (A) and (B) show the mean absolute error of the normalized pressure ($P/T^4$) and energy density ($E/T^4$) as functions of the temperature. The critical temperature $T_c = 0.155$ GeV.} 
    \label{fig:eos}
\end{figure}

Fig. \ref{fig:eos} shows the results for both expanded and unexpanded input dimensions of the neural network. Here, ``unexpanded" refers to using $T$ as the input, while ``expanded" refers to using $(T, T_c)$ as the input, with $T_c = 0.155$ GeV. We can see that the expanded input, labeled as $s(T,T_c)$, demonstrates a significant improvement over the unexpanded, particularly at lower temperatures.

The results indicate that incorporating additional input dimensions significantly enhances the accuracy of the model. This improvement underscores the importance of a more detailed representation of the input space in capturing the complex behavior of QCD matter.

\textbf{Physics-Driven tasks --- Solving Partial Differential Equations.}
We evaluated the impact of input dimension expansion on solving various PDEs using PINNs. The experiments were conducted on a comprehensive set of 20 PDEs. For each PDE, we expanded the input dimensions by including additional constant values derived from the mean of the original input variables. For instance, for a one-dimensional problem with inputs $[x, t]$, we expanded it to $[x, x_c, x, x_c, x, t, t_c, t, t_c, t]$, where $x_c$ and $t_c$ are the mean values of $x$ and $t$, respectively.

\begin{figure}[t]
    \centering
    \subfigure
    {
       \centering
       \includegraphics[width=0.45\columnwidth]{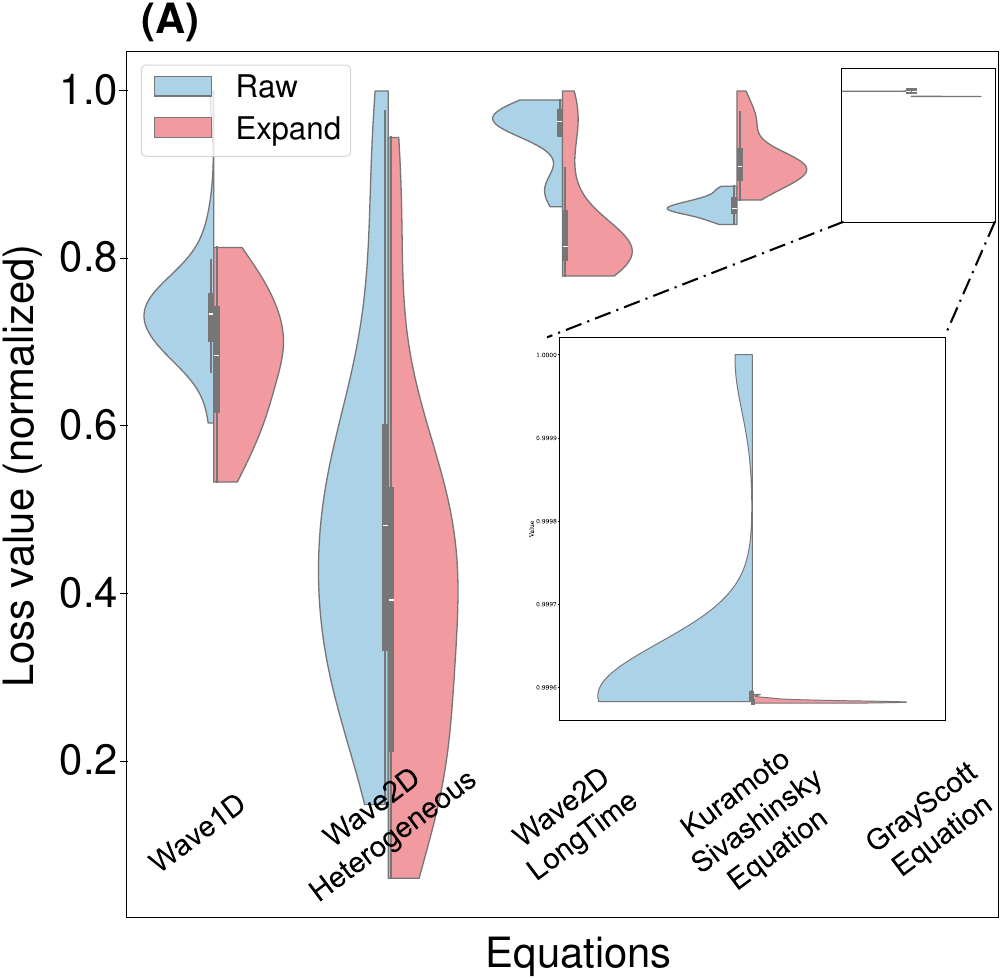}
    }
    \subfigure
    {
       \centering
       \includegraphics[height=0.44\columnwidth]{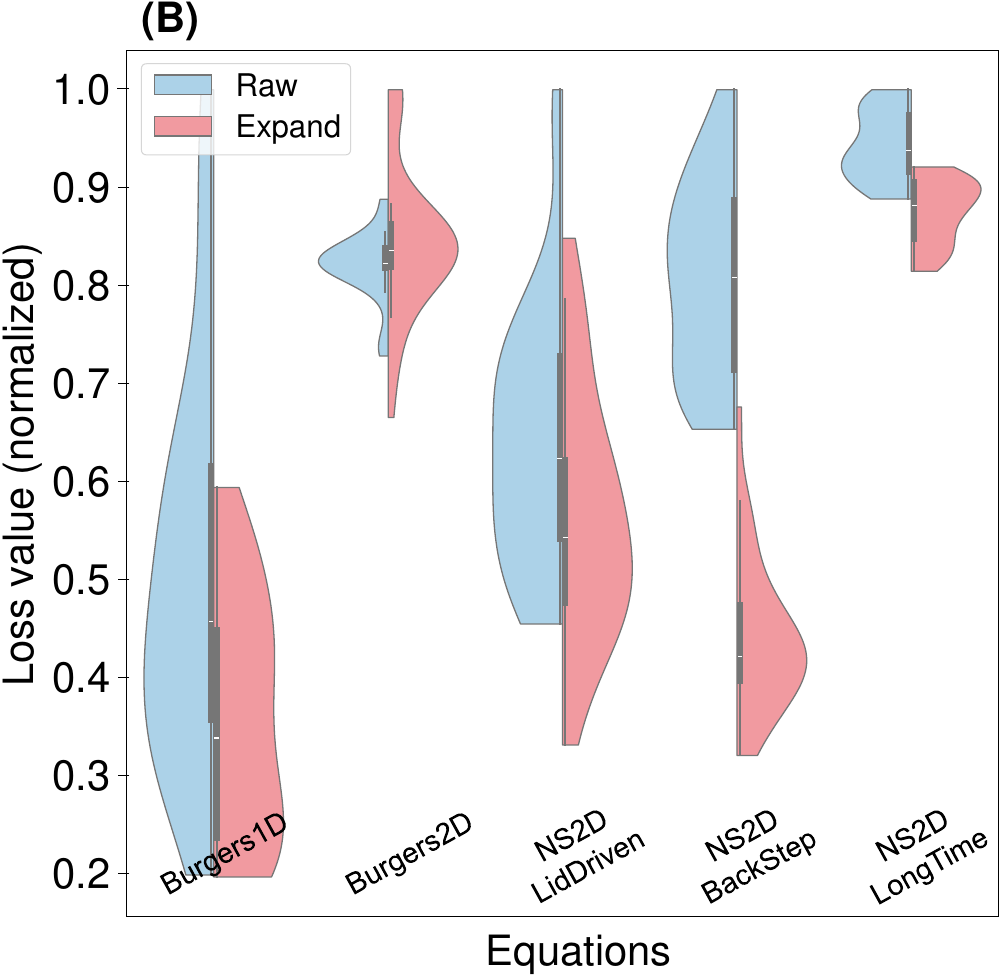}
    }
    \subfigure
    {
       \centering
       \includegraphics[height=0.45\columnwidth]{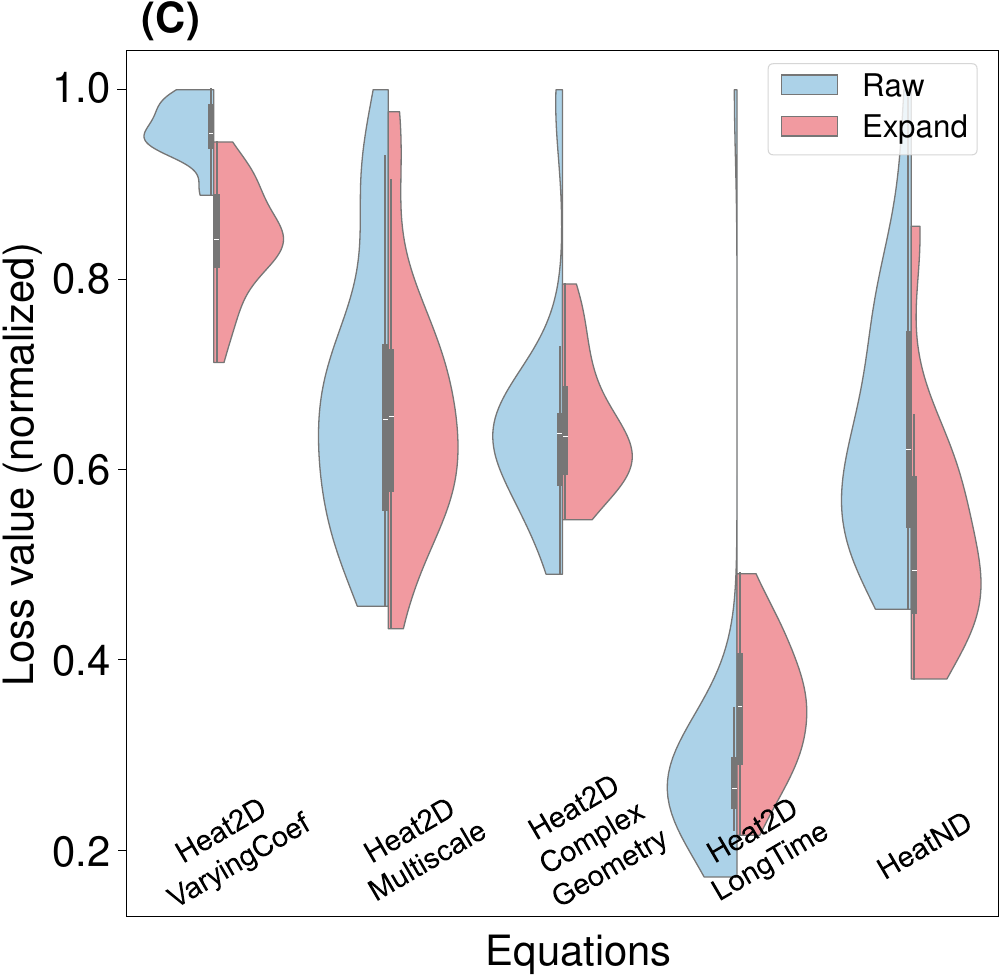}
    }
    \caption{Comparison of loss distributions for 15 PDEs using raw and expanded input dimensions. Each subfigure represents the results for 5 different PDEs. The loss values have been normalized using the maximum loss for each PDE to facilitate comparison. The blue distribution on the left side of each violin plot represents the raw input dimensions, while the light red distribution on the right side represents the expanded input dimensions. The plots show the distribution of loss values across different random seeds, with the expanded input dimensions generally resulting in lower loss values, indicating improved performance.}
    \label{fig-pinns}
\end{figure}

To analyze the performance of the input expansion method, we compared our results with the benchmark provided by the PINNacle framework \cite{hao2023pinnacle}, as shown in Fig. \ref{fig-pinns}. The improvement was evidenced by the distribution of the loss values, where the expanded input dimensions resulted in a lower overall loss compared to the raw input dimensions. Our findings revealed that approximately 75\% of the PDEs showed improved performance with the expanded input dimensions. This improvement was particularly significant for complex and high-dimensional PDEs, demonstrating the effectiveness of input dimension expansion in enhancing the accuracy and robustness of PINNs (the remaining 5 PDEs showed no significant improvement, so they were not included in Fig. \ref{fig-pinns}, which will be explained towards latter Sec. 5.2).

\subsection{Impact of Dimension Expansion on Other Tasks} 
\textbf{Image Coloring.} In the task of image coloring, where the goal is to colorize black-and-white images, we used a generative model based on conditional diffusion to restore grayscale images to color RGB maps with a human-face-based CelebA dataset \cite{liu2015faceattributes}. By both expanding the input dimension of the noisy image, which is used to denoise as a color image, and the grayscale image, which is used as the prompt of the diffusion, we observed a modest improvement in performance. Specifically, the mean absolute error (MAE) of the colorization model was reduced from $8.7\times10^{-3}$ to $8.3\times10^{-3}$. This result demonstrates the effectiveness of input dimension expansion in enhancing neural network performance in image coloring.

\textbf{Sentiment Analysis.} For sentiment analysis tasks using the IMDB dataset \cite{IMDB2011}, we explored the impact of expanding input dimensions on model performance. We modified a BERT-based model \cite{devlin2018bert} by doubling the embedding dimension through an additional linear layer, then mapping it back to the original size to maintain compatibility with the Transformer architecture. This approach allowed us to test the effect of increased input dimensions without altering the overall model structure.

The results showed a modest improvement in accuracy, increasing from 83.40\% to 83.67\%, highlighting the potential benefits of input dimension expansion in enhancing the performance of sentiment analysis models.

\section{Mechanism: Exploring the Symmetry Breaking Hypothesis in Neural Networks}\label{subsec2.2}

\subsection{Symmetry Breaking Mechanism in the Ising Model}

\begin{figure}
    \centering
    \includegraphics[width=1.0\linewidth]{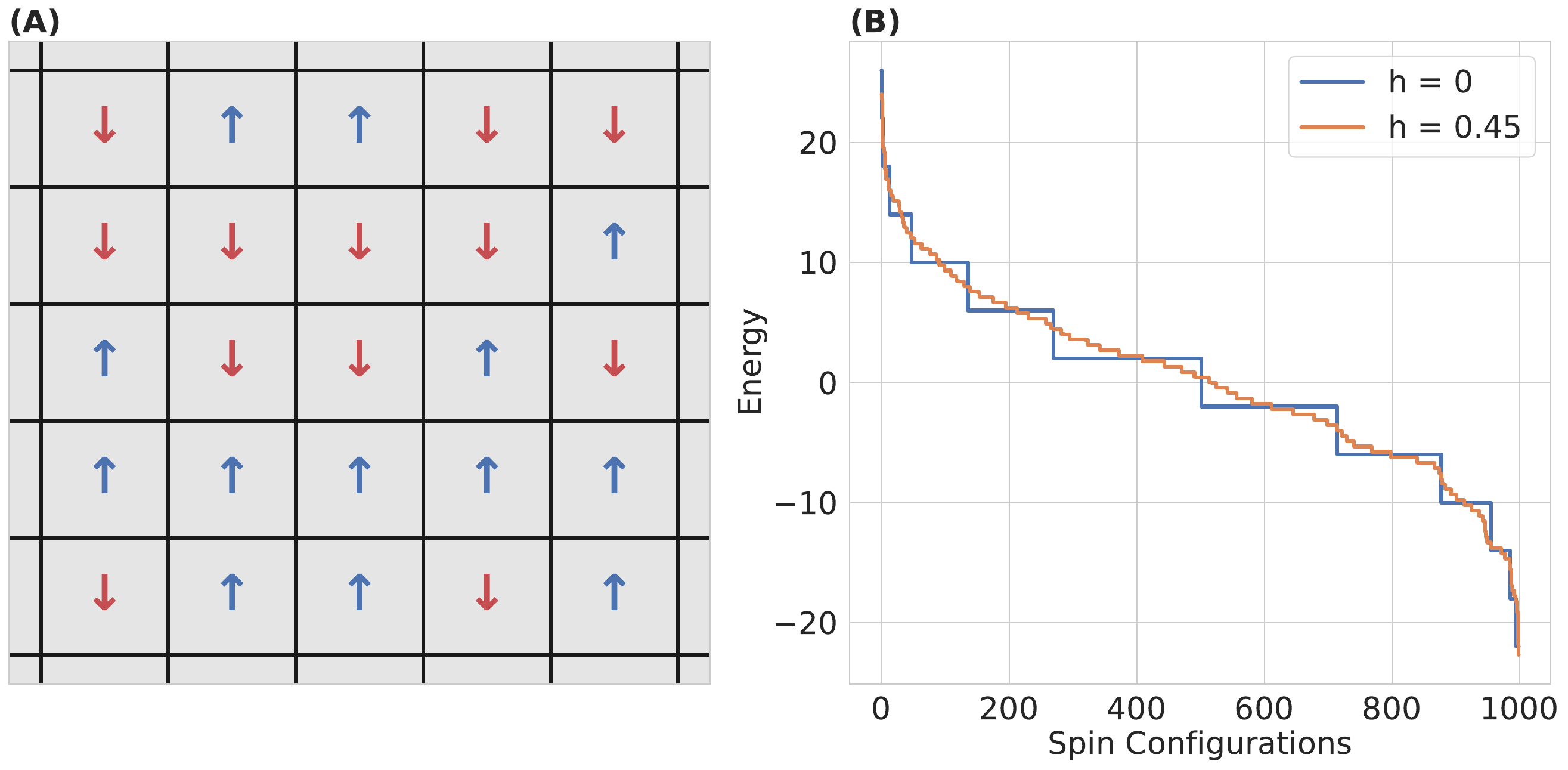}
    \caption{Ising model. (A) Schematic diagram of the Ising model composed of a two-dimensional periodic lattice, typically arranged in a square grid. Each lattice point represents a spin variable, where spin-up is denoted by +1 and spin-down by -1. (B) The energy landscape of the Ising model. The two curves represent the energy landscape for $h=0$ and $h=0.45$. When $h=0.45$, the symmetry of the system is broken.}
    \label{fig:Ising model}
\end{figure}

In physics, symmetry breaking \cite{symmetry1972} is a fundamental concept that elucidates how systems initially exhibiting symmetry can lose this symmetry due to changes in external conditions or internal interactions. Here, we delve into the two-dimensional Ising model \cite{ising1925beitrag}, a widely utilized stochastic process model in physics, primarily employed to study phase transitions and spin-spin interactions. The Ising model is defined on a two-dimensional periodic lattice, typically arranged in a square grid, as depicted in Fig. \ref{fig:Ising model} (A). Each lattice point is assigned a spin variable, which can be either spin-up (denoted by +1) or spin-down (denoted by -1).

In the absence of an external magnetic field, the Ising model exhibits what is termed ``energetic symmetry". This implies that different configurations of spins (i.e., various arrangements of +1 and -1 spins on the lattice) can result in the same total energy for the system. This is because the system's energy $E$ of a configuration is determined by the interactions between neighboring spins, mathematically expressed as:
\begin{eqnarray}
E & = & -J \sum_{\langle i,j \rangle} s_i s_j,
\end{eqnarray}
where $J$ is the interaction strength between neighboring spins, $s_i$ and $s_j$ are the spin values at lattice points $i$ and $j$, and the sum is over all pairs of neighboring spins. In this equation, since the value of $s_i s_j$ remains unchanged when all spins are flipped (changing +1 to -1 and vice versa), the system's total energy remains constant, reflecting the system's symmetry.

However, when an external magnetic field $h$ is introduced, this energy symmetry is disrupted. The interaction between the field and the spins adds a new term to the energy equation:
\begin{eqnarray}
E = -J \sum_{\langle i,j \rangle} s_i s_j - h \sum_i s_i.
\end{eqnarray}

The second term $-h \sum_i s_i$ represents the interaction between the spins and the external magnetic field. This term depends on the orientation of each spin relative to the field: spins aligned with the field (e.g., +1 if $h$ is positive) will lower the energy, while spins opposed to the field will increase the energy. As a result, the previously equivalent energy states of different spin configurations are now differentiated by their alignment with the field, resulting in symmetry breaking.

To illustrate, consider a simple $5\times5$ lattice. Without an external field, the energy of the configuration where all spins are up (+1) is the same as the configuration where all spins are down (-1). However, with a positive external magnetic field, the energy of the all-up configuration will be lower than the all-down configuration, breaking the symmetry. This is depicted in Fig. \ref{fig:Ising model} (B), where the energy landscape changes when $h=0.45$, showing the breaking of symmetry. Since the number of the total spin configurations is $2^{25}$ which is too large to compute, we randomly sampled 1,000 configurations for presentation.

Understanding symmetry breaking in the Ising model provides a foundation for exploring similar concepts in neural networks, where symmetry breaking emerges as a crucial factor in optimization and learning processes.

\subsection{Symmetry Breaking in Neural Networks}

Analogous to the Ising model, consider a neural network with input $x$, weight $w$, and activation function $\sigma$. The output of a two-layer network can be simplified as:
\begin{eqnarray}
\sigma(\sigma(w_{i,j} x_{i})w^{\prime}_{j,1})  & \sim & w_{i,j} x_{i}w^{\prime}_{j,1} +\mathcal{O}(\text{higher order terms)}, \label{eq:output_layers}
\end{eqnarray}
where the same indices are summed. Up to the first order, the output of the above layers is symmetric under the simultaneous change of $w_{i,j_\alpha} \leftrightarrow w_{i,j_\beta}$ and $w^{\prime}_{j_\alpha,1} \leftrightarrow w^{\prime}_{j_\beta,1}$, a property that generates many local minima in the parameter space.

When adding a new dimension $x_{\text{c}}$ to the input, where the subscript $\text{c}$ denotes that the added value is always a constant, Eq. (\ref{eq:output_layers}) changes to:
\begin{eqnarray}
\sigma(\sigma(w_{i,j} x_{i}+w^{\prime\prime}_{1,j} x_{\text{c}})w^{\prime}_{j,1})  & \sim & w_{i,j} x_{i}w^{\prime}_{j,1}+w^{\prime\prime}_{1,j} x_{\text{c}}w^{\prime}_{j,1} 
\nonumber \\
&  &+\mathcal{O}(\text{higher order terms)}, \label{eq:output_2_layers}
\end{eqnarray}
where a new weight $w^{\prime\prime}_{1,j}$ is introduced to fit the new dimension. With the extra term $w^{\prime\prime}_{1,j} x_{\text{c}}w^{\prime}_{j,1}$, the aforementioned symmetry property is broken, decreasing the number of local minima similar to the Ising model.

To demonstrate that the additional dimension in the input breaks the symmetry, we perform an example of number classification with a neural network whose parameters take values of either -1 or 1. A random number between 0 and 1 is chosen and classified as category ``0" if it is smaller than 0.5, and as category ``1" otherwise. We considered two scenarios with two four-layer neural networks.

\begin{figure}[h]
    \centering
    \includegraphics[width=1\textwidth]{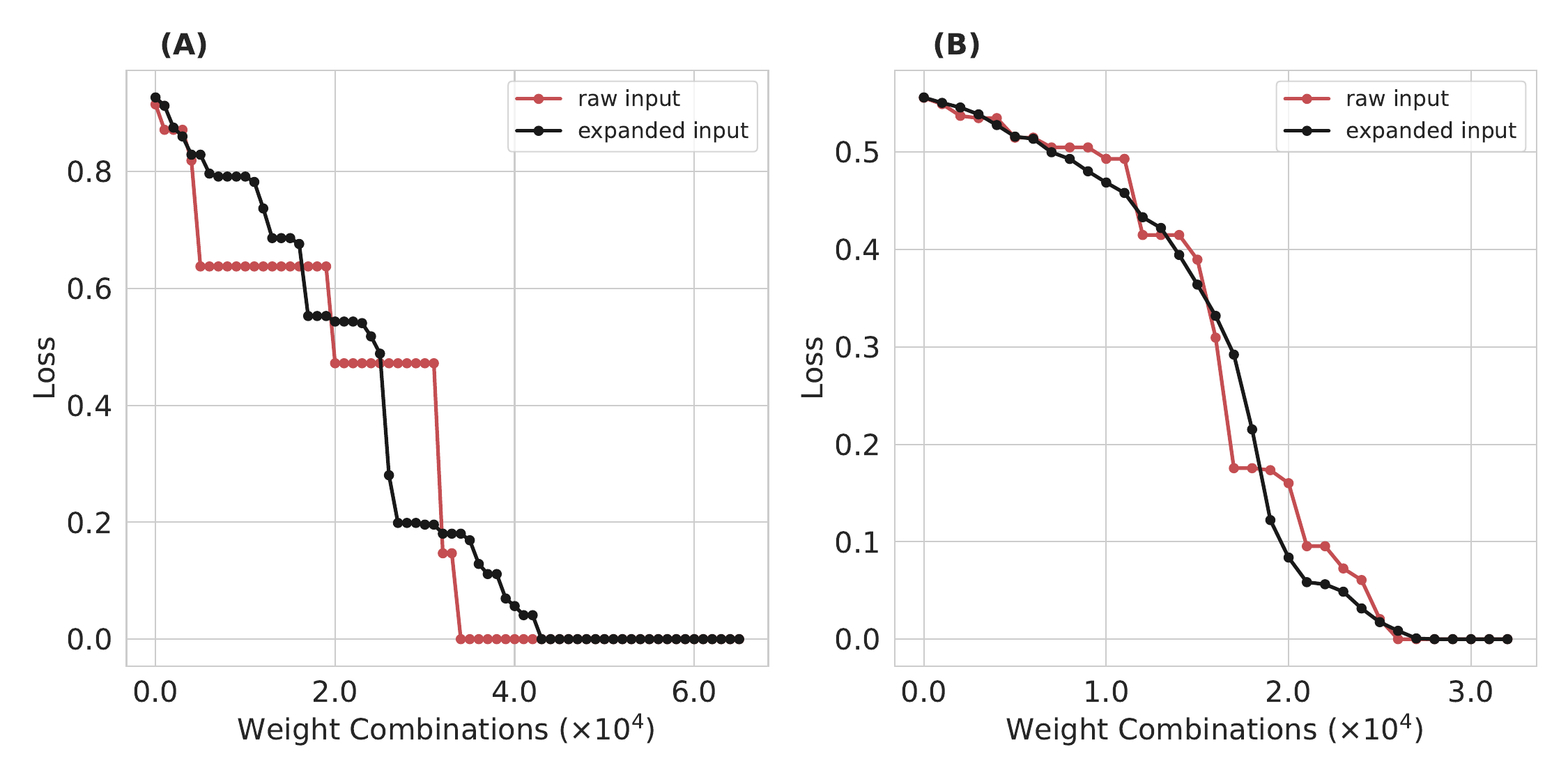}
    \caption{Comparison of MSE Loss values with different weights for binary classification (random number). The horizontal axis represents different weight combinations, and the vertical axis represents the MSE loss. The MSE loss values are sorted in descending order before plotting, so the weight combinations on the horizontal axis do not correspond to the same weights across different scenarios. Subfigures (A) and (B) correspond to comparisons without and with the bias in the linear layers, respectively.}
    \label{fig:loss landscape expansion}
\end{figure}

The influence of incorporating an additional dimension is illustrated in Fig. \ref{fig:loss landscape expansion} (A). This figure demonstrates that the introduction of the additional dimension leads to a reduction in the number of degenerate states—defined as distinct parameter configurations that produce identical loss values. Consequently, this results in a smoother training trajectory and diminishes the probability of the network becoming trapped in local minima. A similar phenomenon is also observed when the network contains a bias vector in the linear layers (Fig. \ref{fig:loss landscape expansion} (B)).

Thus, the introduction of an additional input dimension can break the symmetry in neural networks, similar to how an external magnetic field breaks the symmetry in the Ising model. This symmetry breaking can lead to a smoother training process and reduce the likelihood of the network getting stuck in local minima, thereby improving the overall performance of the neural network.

\section{Evidence and Metric: Symmetry Breaking Techniques and Its Measurement}

\subsection{Symmetry Breaking Effects of Equivariance, Dropout, and Batch Normalization}

To investigate the symmetry-breaking effects of equivariance \cite{cohen2016equivariant}, dropout \cite{srivastava2014dropout}, and batch normalization \cite{ioffe2015batchnormalization} in neural networks, we conducted a series of controlled experiments using a simple CNN architecture. We compared the performance of five different network configurations: a baseline CNN, a CNN with dropout, a CNN with batch normalization, a CNN with equivariance constraints, and a CNN with incorrect symmetry embedded in the structure (wrong equivariance).

We constructed a synthetic dataset consisting of 12 binary images, each of size 
2×2 pixels, with corresponding binary labels. The images were designed to exhibit simple patterns with pixel values of -1, 0, or 1, and are invariant under 180-degree rotations and flips but not under 90-degree rotations --- a property crucial for evaluating the effects of equivariance in neural networks. The labels were assigned based on the presence of specific patterns in the images, creating a binary classification task.

We generated all possible combinations of weights for the convolutional and fully connected layers, with each weight taking values of either -1 or 1. This exhaustive search allowed us to evaluate the loss landscape comprehensively.

\begin{figure}[h]
\centering
\includegraphics[width=0.8\textwidth]{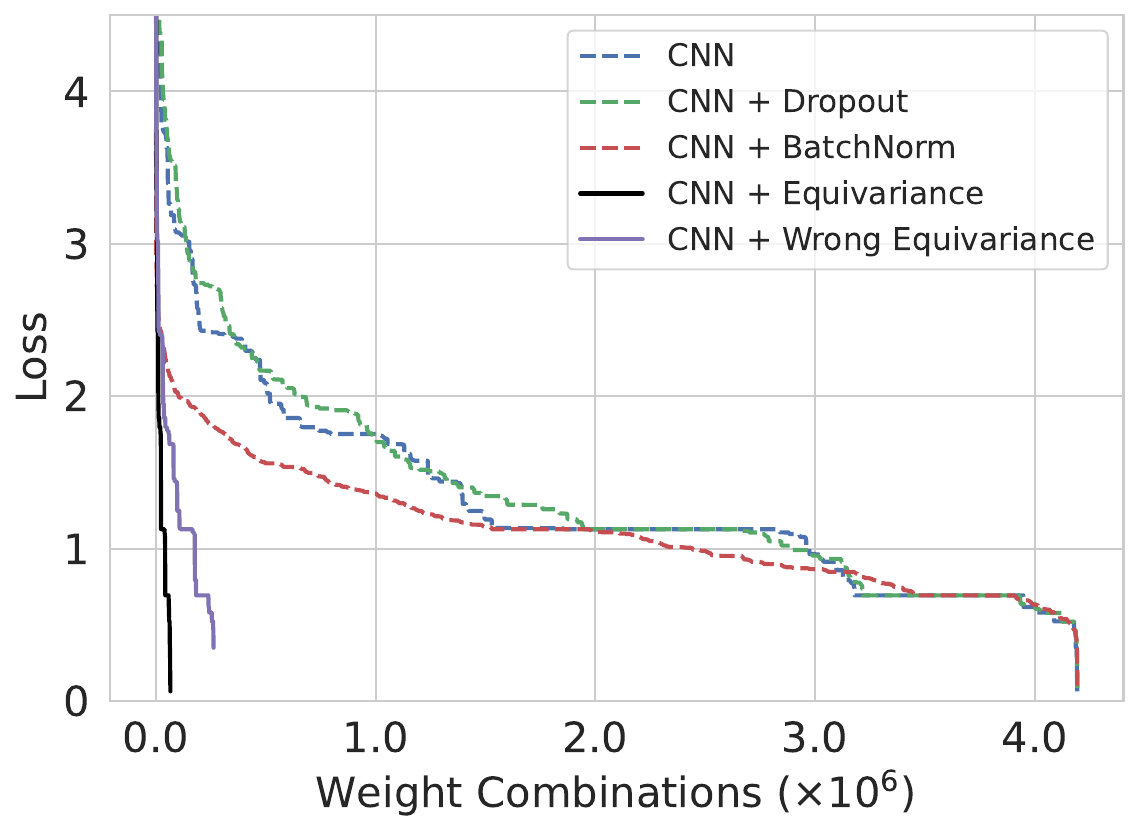}
\caption{Loss values with different weights of binary classification (four pixel images) for cases: a baseline CNN, a CNN with dropout, a CNN with batch normalization, a CNN with equivariance constraints, and a CNN with wrong equivariance. The horizontal axis represents different weight combinations, where each number corresponds to a specific set of weights. The vertical axis represents the loss values, which are sorted in descending order before plotting, so the weight combinations on the horizontal axis do not correspond to the same weights across different scenarios. The Baseline ConvNet shows a plateau region (x-axis between 1.5 and 2.9) where the loss remains constant despite different weight combinations, indicating a symmetry in these weight combinations from the perspective of loss values. Adding dropout and batch normalization disrupts this plateau in the loss landscape, helping to avoid optimization getting stuck in these flat regions. For the cases with equivariance constraints, the number of weight combinations is reduced compared to other scenarios. This reduction is due to the introduction of equivariance, which eliminates the need to consider many redundant weight combinations. As a result, the curves for the equivariant and wrong equivariant constraints are shorter along the horizontal axis. Both equivariant and wrong equivariant constraints also break the degeneracy by reducing the number of states. Additionally, the wrong equivariant constraint fails to converge to the minimum loss value.}
\label{fig:loss landscape equi-batch-drop}
\end{figure}

To assess the symmetry-breaking effects, we compared the loss values across the different network configurations. The results of our experiments are summarized in Fig. \ref{fig:loss landscape equi-batch-drop}. The figure shows the sorted loss values for each network configuration, highlighting the impact of dropout, batch normalization, and equivariance on the loss landscape.

Our findings indicate that there are two primary approaches to break such symmetry: splitting the degenerated states as seen in dropout, batch normalization, or reducing the number of parameter combinations, as seen in the equivariance cases.

Among these methods, equivariance stands out as a more efficient approach compared with dropout and batch normalization. By embedding symmetry constraints directly into the network architecture, equivariance reduces the complexity of the parameter space, leading to a more streamlined and effective training process. This observation suggests that incorporating symmetry principles into network design could be a promising direction for future research, potentially offering advantages over solely applying various regularization techniques or increasing network size.

Meanwhile, our experiments also reveal a critical insight: embedding incorrect symmetry into the network can prevent convergence to the optimal solution. This underscores the importance of identifying and embedding the correct symmetries, which requires a deep understanding of the underlying data and problem domain.

\subsection{Measuring the Degree of Symmetry Breaking}

To quantify the degree of symmetry breaking in neural networks, we developed a metric called the ``Replica Distance (RD) Metric". This metric evaluates the Wasserstein distance between different weight configurations of the network (i.e., replicas), providing insights into the extent of symmetry breaking. We applied the RD Metric to the CIFAR-10 dataset using five distinct neural network architectures: SimpleCNN, DropoutCNN, BatchNormCNN, FlipEquivarianceCNN, and RotationEquivarianceCNN (180 degrees), analogous to that in Fig. \ref{fig:loss landscape equi-batch-drop}. Here, it is important to note that the CIFAR-10 dataset does not exhibit 180-degree rotational equivariance, which makes the rotation constraint less appropriate.

\begin{figure}[h]
\centering
\includegraphics[width=1\textwidth]{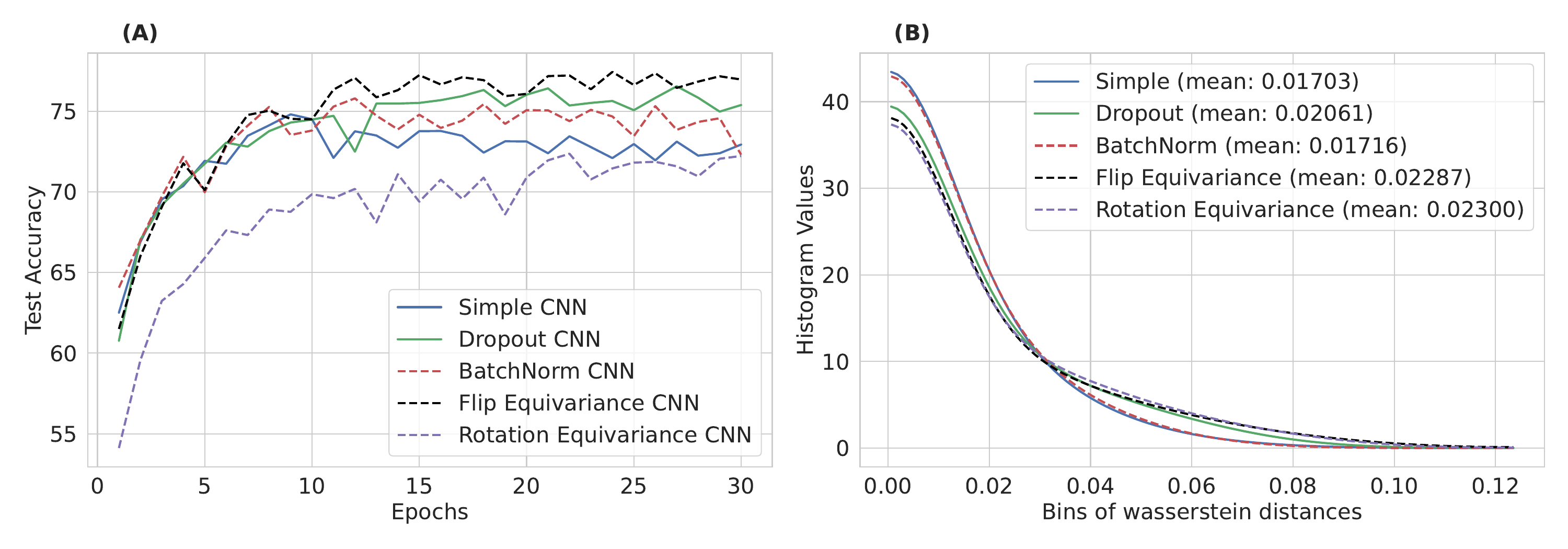}
\caption{(A) Test accuracies for different neural network architectures on the CIFAR-10 dataset. (B) Histogram of Wasserstein distances between weight distributions of trained models. The peak position (value at the x-axis) of the histogram serves as a measure of the degree of symmetry breaking.}
\label{fig:training_results}
\end{figure}

After training the test accuracies were recorded for each epoch and are shown in Fig. \ref{fig:training_results} (A), where we see that the FlipEquivarianceCNN achieved the highest test accuracy, followed by DropoutCNN, BatchNormCNN, SimpleCNN, and RotationEquivarianceCNN.

To measure the degree of symmetry breaking, we used a subset of 100 images from the CIFAR-10 dataset, a scenario that is common in network design because we do not want to use the entire dataset during the design and selection phase. Analogous to the ``Parassi" scheme used to determine the replica symmetry of the spin-glass system \cite{parasi1988, liao2024losslandscapeslens}, we trained each model for 200 epochs under 200 different random seeds, and the final weights were cached. Thus, each model has 200 different weight combinations.

The degree of symmetry breaking was then quantified by calculating the Wasserstein distance between the weight distributions of the trained models. The Wasserstein distance is a measure of the distance between two probability distributions, which in this context helps us understand how different the weight distributions are after training under different initial conditions. A larger Wasserstein distance indicates greater variability in the weight configurations, which corresponds to a higher degree of symmetry breaking.

For each model, we paired these 200 different weight combinations and calculated a series of Wasserstein distances, then plotted histograms of these distances and applied Gaussian smoothing to obtain smooth curves. These smoothed histograms were normalized to form probability density functions. The five models correspond to five curves (Fig. \ref{fig:training_results} (B)). The mean value of the probability density was used as the metric. A higher average distance indicates a greater degree of symmetry breaking. The FlipEquivarianceCNN exhibited the highest degree of symmetry breaking, followed by RotationEquivarianceCNN, DropoutCNN, BatchNormCNN, and SimpleCNN.

Interestingly, despite its high degree of symmetry breaking, the RotationEquivarianceCNN showed the lowest test accuracy. This can be attributed to the fact that the CIFAR-10 dataset does not exhibit 180-degree rotational equivariance, making the rotation equivariance constraint inappropriate. In contrast, the FlipEquivarianceCNN, which leverages horizontal flip equivariance, achieved the highest test accuracy, demonstrating the effectiveness of appropriate equivariant constraints.

Our proposed metric suggests a way to measure the degree of symmetry of neural networks. By analyzing the Wasserstein distance between weight distributions, we can quantify the extent to which different techniques, such as input dimension expansion, dropout, batch normalization, and equivariance constraints, break symmetry in the parameter space. This metric can guide the design of neural network architectures and optimization strategies, helping to improve model performance and generalization.

\section{Discussion and Conclusion}\label{sec12}

In this study, we explored the principle of symmetry breaking in neural network optimization through three key findings: input dimension expansion, unified optimization techniques, and the development of a symmetry breaking metric. Here, we summarize our key findings, discuss their implications, and outline future research directions.

\subsection{Key Findings}

\textbf{Input Dimension Expansion.} Our experiments demonstrated that expanding input dimensions with constant values significantly improves performance across various tasks, including image classification, Physics-Informed Neural Networks (PINNs), image coloring, and sentiment analysis. This technique leverages symmetry breaking to facilitate smoother optimization and better generalization by providing additional degrees of freedom for the model to explore during training. However, it is crucial to recognize that input expansion is not a universal solution and has its limitations. Excessively increasing input dimensions can lead to inefficiencies, and the effectiveness of input expansion can vary depending on the initial network design and the specific task.

\textbf{Symmetry Breaking as a fundamental optimization principle.} We drew an analogy between the symmetry breaking mechanism in the Ising model and neural networks. By introducing additional input dimensions, we break the inherent symmetries in the network, leading to a reduction in the number of degenerate states and a smoother training trajectory. This symmetry breaking mechanism helps the network escape local minima and saddle points in the loss landscape, thereby facilitating better optimization.

\textbf{Symmetry Breaking Techniques and Measurement.} We investigated the symmetry-breaking effects of equivariance, dropout, and batch normalization in neural networks. Our findings indicate that these techniques effectively break the symmetry in the loss landscape, leading to more robust optimization. Among these methods, embedding equivariance constraints directly into the network architecture proved to be particularly effective. Additionally, we developed a metric to quantify the degree of symmetry breaking in neural networks. By analyzing the Wasserstein distance between weight distributions, we can measure the extent of symmetry breaking.

\subsection{Challenges and Future Directions}

While our results are promising, further research is still essential to further validate and refine the symmetry breaking metric across different datasets and neural network architectures. Extending the analysis to a broader range of datasets, including CIFAR-100, ImageNet, and other domain-specific datasets, will help us more sufficiently and deeply understand how the symmetry metric performs across various types of data and tasks, providing a more comprehensive evaluation of its effectiveness. In fact, our experiments have shown that the worse the original network performs, the more significant the improvement brought by input dimension expansion. Therefore, for some tasks where the network design is already very good, the effect of input dimension expansion is not obvious, or even tends to certainly degenerate the performance due to the good symmetry breaking properties already possessed by the original network architectures, such as in several cases of PDE solving, sentiment analysis and image coloring tasks.

Refining the symmetry metric itself is also a critical area for future work. Our proposed metric relies on the choice of several parameters and could be unstable depending on the random seeds. Hence, exploring alternative and comprehensive formulations of the metric will capture more aspects of the loss landscape, offering more nuanced insights into symmetry breaking. By developing a more sophisticated and accurate metric, we can better quantify the degree of symmetry breaking in neural networks and understand its impact on optimization and performance.

Moreover, investigating the practical applications of the symmetry metric is beneficial for translating our theoretical findings into real-world benefits. By using the symmetry metric to guide the design of neural network architectures and optimization strategies, we can develop more efficient and effective models. This approach can help identify the most appropriate symmetries to embed in network designs, leading to improved performance and generalization.

Another promising direction is the use of neural networks themselves to identify equivariance within the data. By training networks to recognize and exploit these properties, we can develop models that are inherently more efficient and effective. This approach could be particularly valuable for scientific and engineering applications, where the tasks often involve complex and domain-specific equivariance. By identifying and embedding the equivariance, we can improve the performance of neural networks even in scenarios with limited data and computational resources.

\subsection{Conclusion}

In conclusion, our exploration of the principle of symmetry breaking in neural network optimization has yielded  insights and practical techniques for enhancing model performance. By understanding and leveraging symmetry breaking, we can develop more efficient and effective AI systems, thereby contributing to the advancement of both AI and its interdisciplinary applications. Continued research in this area holds the potential to further improve our understanding and optimization of neural networks, paving the way for more interpretable AI models.

Our findings highlight the value of interdisciplinary approaches, demonstrating how principles from physics can inform and advance AI research.

\section{Methods}\label{sec11}

\subsection{Methodology for Expanding Input Dimensions in Image Classification}

\begin{figure}
    \centering
    \includegraphics[width=1.0\linewidth]{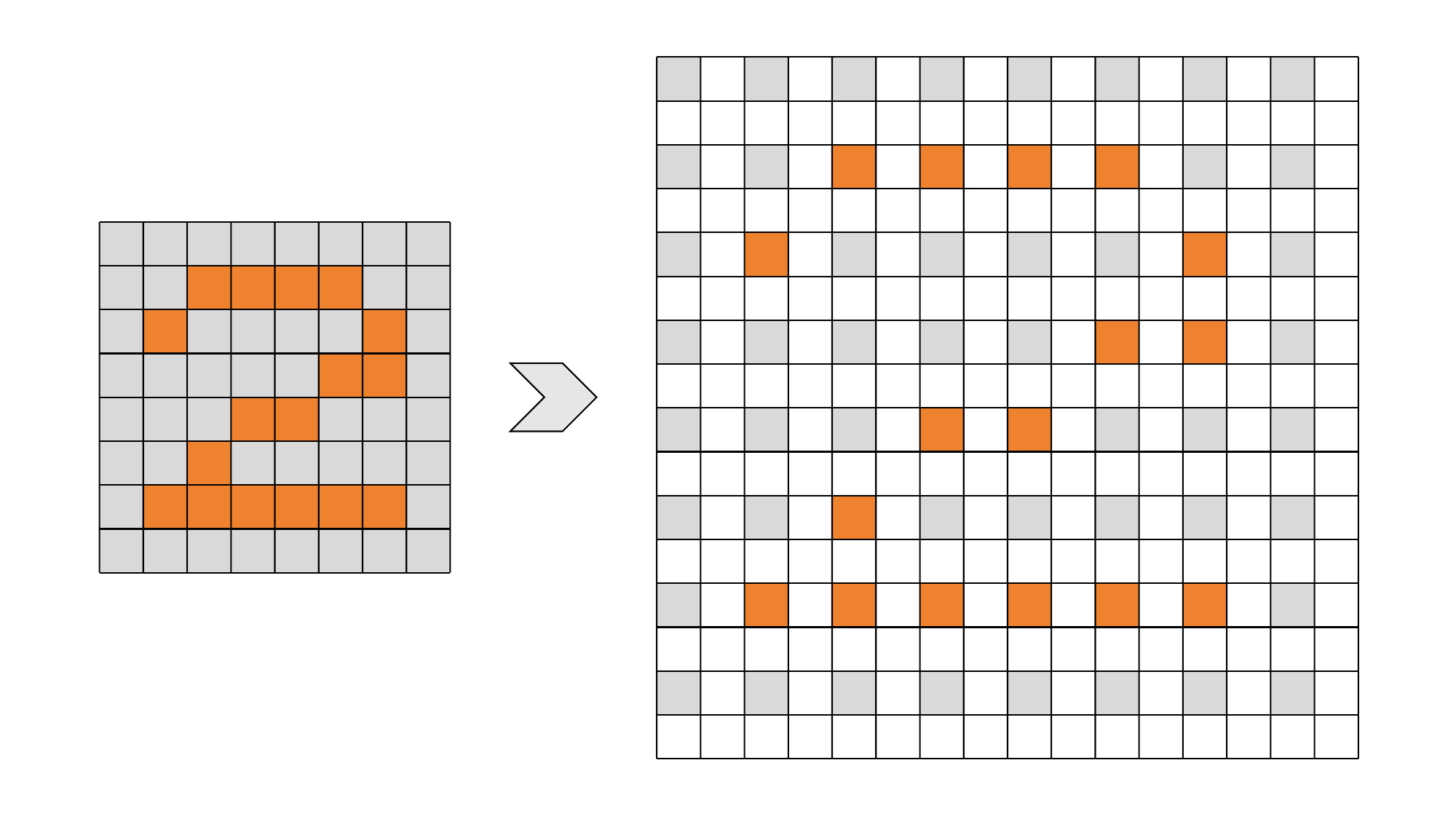}
    \caption{Illustration of image pixel expansion. The figure exemplifies the expansion method for the digit "2", where the expansion factor is 2, i.e., $K=2$. The blank pixels in the expanded image are filled by constant values (e.g., 0.5).}
    \label{fig:expansion}
\end{figure}

In identify the symmetry breaking principle, we coincidentally found a method that enhances image classification performance by expanding the spatial dimensions of input images. This approach involves filling the expanded positions with a fixed value, followed by a careful design of the neural network architecture to manage the increased dimensionality. For instance, if the original image is $N \times N$ pixels, we expand it by a factor of $K$ as an image with $NK \times NK$ pixels. In this $NK \times NK$ image, for example, the original pixels of $[0,0], [0,1], [0,2], \cdots, [1,0], [1,1], \cdots, [N, N]$ are positioned at coordinates as $[0,0], [0,K], [0,2K], \cdots, [K,0], [K,2K], \cdots, [N(K-1), N(K-1)]$, while the remaining positions are filled with the value 0.5 (an example is shown in Fig. \ref{fig:expansion}). This method ensures that the spatial expansion maintains the relative positions of the original pixels while introducing new elements with a uniform value to fill the increased space.

In our experiments, we applied this expansion method with a factor of $K=2$ for all datasets and neural network models. This expansion technique allows us to maintain a consistent number of neural network parameters by using a pooling method on the feature map extracted by the last convolutional layer. Consequently, the size of the feature map output depends only on the number of channels, not the input data size. During training, we employed the Stochastic Gradient Descendent optimizer with an initial learning rate of 0.1, momentum of 0.9, and a weight decay of $5\times10^{-4}$ for all neural network models. For the CIFAR, KMNIST, EuroSAT, and other datasets except for ImageNet, we used the CosineAnnealing method to adjust the learning rate, with $T_{max}$ set to match the total number of training epochs. For the ImageNet-R and ImageNet-100 datasets, we decayed the learning rate by 0.2 at epochs 90, 120, and 160. This approach ensures that performance improvements are attributed to the image expansion technique rather than an increase in network parameters.

\subsection{Methodology for Expanding Input Dimensions in AI4Science}

\subsubsection{Quantum Chromodynamics}

In QCD, the partition function $\ln Z(T)$ is fundamental for calculating the pressure $P(T)$ and energy density $\epsilon(T)$ using statistical dynamics formulae,
\begin{align}
\ln Z(T) &= \ln Z_\text{g}(T) + \ln Z_{\text{u/d}}(T) + \ln Z_\text{s}(T), 
\nonumber\\
P(T) &= T \left( \frac{\partial \ln Z(T)}{\partial V} \right)_{T}, 
\nonumber\\
\epsilon(T) &= \frac{T^{2}}{V} \left( \frac{\partial \ln Z(T)}{\partial T} \right)_{V},
\label{eq:lnz_total}
\end{align}
where $\ln Z_\text{g}(T)$, $\ln Z_{\text{u/d}}(T)$, and $\ln Z_\text{s}(T)$ are the partition functions for gluons, u/d quarks, and s quarks, respectively. These partition functions can be computed from the momentum integration,
\begin{align}
\ln Z_\text{g}(T) &= - \frac{d_\text{g} V}{2 \pi^{2}} \int_{0}^{\infty} p^{2} dp \ln \left[ 1 - \exp \left( -\frac{1}{T} \sqrt{p^{2} + m_\text{g}^{2}(T)} \right) \right], 
\nonumber\\
\ln Z_{q_i}(T) &= + \frac{d_{q_i} V}{2 \pi^{2}} \int_{0}^{\infty} p^{2} dp \ln \left[ 1 + \exp \left( -\frac{1}{T} \sqrt{p^{2} + m_{q_i}^{2}(T)} \right) \right].
\label{eq:lnz_func}
\end{align}
Here, $d_\text{g} = 16$, $d_{q_\text{s}} = 12$, and $d_{q_{\text{u/d}}} = 24$ are the degrees of freedom for gluons and quarks. These integrals can be numerically evaluated using a 50-point Gaussian quadrature method \cite{1969Calculation}. With the calculated $P(T)$ and $\epsilon(T)$, the Eos can be obtained, linking the entropy density $s$ with $P(T)$ and $\epsilon(T)$ as,
\begin{align}
s &= \frac{\epsilon + P}{T}.
\label{eq:eos_spt}
\end{align}

In the neural network implementation, we employ three mass models to represent the quasi-particles. Specifically, we define the masses for u/d quarks, s quarks, and gluons as $m_{\text{u/d}}(T,\theta_1)$, $m_{\text{s}}(T,\theta_2)$, and $m_{\text{g}}(T,\theta_3)$, where $\theta_1$, $\theta_2$, and $\theta_3$ are the network parameters. Each mass model consists of 7 residual blocks with 32 neurons and sigmoid activation functions. The input is the temperature $T$, and the output is the corresponding quasi-quark and quasi-gluon mass. Once the masses are determined, they are substituted into Eq. (\ref{eq:lnz_func}), and the 50-point Gaussian quadrature method is used to compute the partition function $\ln Z$. The QCD equation of state is then calculated using Eqs. (\ref{eq:lnz_total}) and (\ref{eq:eos_spt}). The training objective is to minimize the mean squared error between the neural network output and the Lattice QCD results from the HotQCD collaboration \cite{HotQCD:2014kol}. For the expanded case, we only modify the input dimension, expanding from $T$ to $(T, T_C = 0.155 \text{ GeV})$.

\subsubsection{Solving Partial Differential Equations}

\textbf{Experimental Setup.} We utilized a simple feedforward neural network (FNN) architecture with five fully connected layers, each containing 100 nodes. The PDEs were selected from the PINNacle benchmark, which includes a diverse range of PDEs such as Burgers' equation, Poisson's equation, heat equation, Navier-Stokes equations, wave equations, and chaotic PDEs like the Gray-Scott and Kuramoto-Sivashinsky equations. The experiments were conducted using the following setup:

\textbf{PDE List.} The list of PDEs included \text{Burgers 1D}, \text{Burgers 2D}, \text{Poisson 2D}, \text{Poisson Boltzmann 2D}, \text{Poisson 3D with Complex Geometry}, \text{Poisson 2D of Many Area}, \text{Heat2D with Varying Coefficient}, \text{Multiscale Heat2D}, \text{Heat2D with ComplexGeometry}, \text{LongTime Heat2D}, \text{NS2D with LidDriven}, \text{NS2D with BackStep}, \text{NS2D with LongTime}, \text{Wave 1D}, \text{Heterogeneous Wave 2D}, \text{LongTime Wave2D}, \text{Kuramoto Sivashinsky Equation}, \text{GrayScott Equation}, \text{Poisson ND}, and \text{Heat ND}.

\textbf{Training Configuration.} The neural network was configured with 5 hidden layers each of 100 nodes and a learning rate of $10^{-3}$. The training was conducted for 20,000 iterations. The experiments were repeated 30 times with different random seeds ranging from 0 to 29 to ensure robustness.

\textbf{Implementation.} The experiments were implemented using the DeepXDE \cite{lu2021deepxde} library with PyTorch backend. The training process used Adam optimizer. The models were trained using a combination of PDE residuals, boundary conditions, and available data losses exactly the same as in PINNacle.

\textbf{Analysis.} To analyze the performance of the input expansion method, we focused on the distribution of the loss values for both the raw and expanded input methods. The loss values were calculated as the MSE between the predicted and true solutions of the PDEs.

\subsection{Methodology for Expanding Input Dimensions in Other Tasks}

\subsubsection{Image Coloring}

In the image coloring task, we use the CelebA dataset, where a generative model based on conditional diffusion is employed to restore grayscale images to color RGB maps, utilizing a U-Net architecture \cite{ronneberger2015u} as the primary backbone. During training, we began with a three-channel color map and, similar to the DDPM approach \cite{ho2020denoising}, progressively added Gaussian random variables, diffusing the image into nearly complete random noise. In the recoloring process, a random matrix is first sampled, and then the grayscale map is input into the U-Net as a condition. The denoising network then gradually removes the noise, generating a color image. Instead of using an implicit diffusion model, we directly diffuse the color map. The network work is trained with $4\times10^5$ steps with a learning rate of $10^{-4}$.

To expand the input, we applied the same pixel-level expansion technique used in image classification to both the color map before diffusion and the grayscale map used as conditional input. This expansion increases the size of the noise matrix that the diffusion model needs to predict. To avoid the additional complexity for the neural network of learning to add noise to the newly filled pixels, we excluded the labels corresponding to these expanded pixel positions when calculating the color loss on the expanded image. This exclusion ensures that the neural network focuses only on the original pixel positions, simplifying the learning process.

An additional advantage of this approach is that it allows for a direct comparison of the MSE between the expanded and unexpanded images. Since the number of pixels considered in the MSE calculation remains the same for both cases, we can effectively assess the impact of input expansion on image coloring performance by comparing the MSE values of the two methods.

\subsubsection{Sentiment Analysis}

We describe the method used to evaluate the effect of expanding input dimensions on sentiment analysis tasks using the IMDB dataset. The goal was to assess whether adding constant values to the input dimensions could improve the performance of a BERT-based model for binary classification.

\textbf{Data Preparation.} We utilized the IMDB dataset, which contains movie reviews labeled as either positive or negative. The dataset was divided into training and testing sets, each containing reviews stored in separate directories for positive and negative sentiments.

\textbf{Data Preprocessing.} We used the \texttt{DistilBertTokenizer} to tokenize the text data. The tokenizer was configured to truncate and pad the sequences to a maximum length of 256 tokens, ensuring that all input sequences are of uniform length, which is necessary for batch processing in neural networks.

\textbf{Model Architecture.} We modified the BERT model to include an expanded embedding layer. The embedding layer was replaced with a custom \texttt{FixedBertEmbedding} class that allowed for the expansion of input dimensions. In the baseline model, we used the standard BERT embedding layer. This model served as the control to compare the effects of input dimension expansion. The expanded embedding model included a linear layer to map the expanded input dimensions back to the original embedding size. This modification aimed to introduce additional features into the input space, potentially aiding the model in capturing more complex patterns in the data to break the proposed symmetry.

The \texttt{FixedBertEmbedding} class was designed to freeze the original BERT embedding layer, preventing updates during training. The original embeddings were expanded by interleaving them with zeros, effectively doubling their dimensions. A linear layer was then used to map these expanded dimensions back to the desired embedding size.

Beyond the embedding layer, the Transformer architecture of the modified BERT model consisted of 4 layers, each with a hidden size of 48. Each Transformer layer included 4 attention heads, allowing the model to focus on different parts of the input sequence simultaneously. The intermediate size for the feed-forward network within each layer was set to 3072, providing a substantial capacity for learning complex representations. The model supported a maximum of 512 position embeddings, accommodating long input sequences. Additionally, the type vocabulary size was set to 2, enabling the use of token type embeddings for distinguishing between different segments of the input.

The input to the model included token IDs, token type IDs, and position IDs, which were processed through the expanded embedding layer. The output from the embedding layer was then fed into the Transformer layers. The final output of the model was a classification logits vector, which was used to predict the class labels. This detailed configuration ensured that the model could effectively handle the expanded input dimensions and leverage the additional features introduced by the modified embedding layer.

\textbf{Training and Evaluation.} The training arguments were configured to save the best model and evaluate the performance at each epoch. Key parameters included a learning rate of $2\times10^{-5}$, a batch size of 32, and a total of 30 training epochs. The evaluation strategy was set to assess the model’s performance at the end of each epoch.

\subsection{Analyzing the Loss Landscape in Random Number Classification}

To investigate the impact of input dimension expansion on the loss landscape of neural networks, we conducted experiments using a simple feedforward neural network for binary classification of random numbers.

\textbf{Input Data.} We used a random number between 0 and 1 as input to the network. The number was labeled as 0 if it was less than or equal to 0.5, and as 1 otherwise.

\textbf{Model Architectures.} To compare cases for raw (the usual input) and expanded (expanded input), we employed two network configurations:

A feedforward neural network with an input layer of size 1, followed by three hidden layers of sizes 3, 3, and 2, and an output layer of size 1. The activation function used was the hyperbolic tangent (tanh).

A similar network but with an expanded input layer of size 2. In this configuration, the original input was augmented with an additional constant value of 0.5. For example, if the original input was 0.7, the expanded input would be the vector [0.7, 0.5].

\textbf{Loss Landscape Analysis.} To evaluate the loss landscapes, we varied the weights of the network layers, allowing each weight to take values of -1 or 1. For each combination of weights, we computed the network's output and the corresponding loss using the MSE loss function. 

To explicitly demonstrate the effect of the additional dimension, we considered two scenarios in our analysis. One did not include bias terms in the network layers, so the loss was computed solely based on the weights and the input data; and another included bias terms in the first layer of the network.

By systematically varying the weights and biases, we were able to visualize and compare the loss landscapes. This approach allowed us to assess how input dimension expansion and the inclusion of bias terms influence the training dynamics and optimization process of neural networks.

\subsection{Analyzing the Loss Landscape of Equivariance, Dropout, and Batch Normalization}

To systematically investigate the effects of equivariance, dropout, and batch normalization on the loss landscape of neural networks, we designed a series of controlled experiments using a simple CNN architecture and a synthetic dataset. The following outlines the methodology employed in our study.

\begin{figure}[h]
    \centering
    \includegraphics[width=\textwidth]{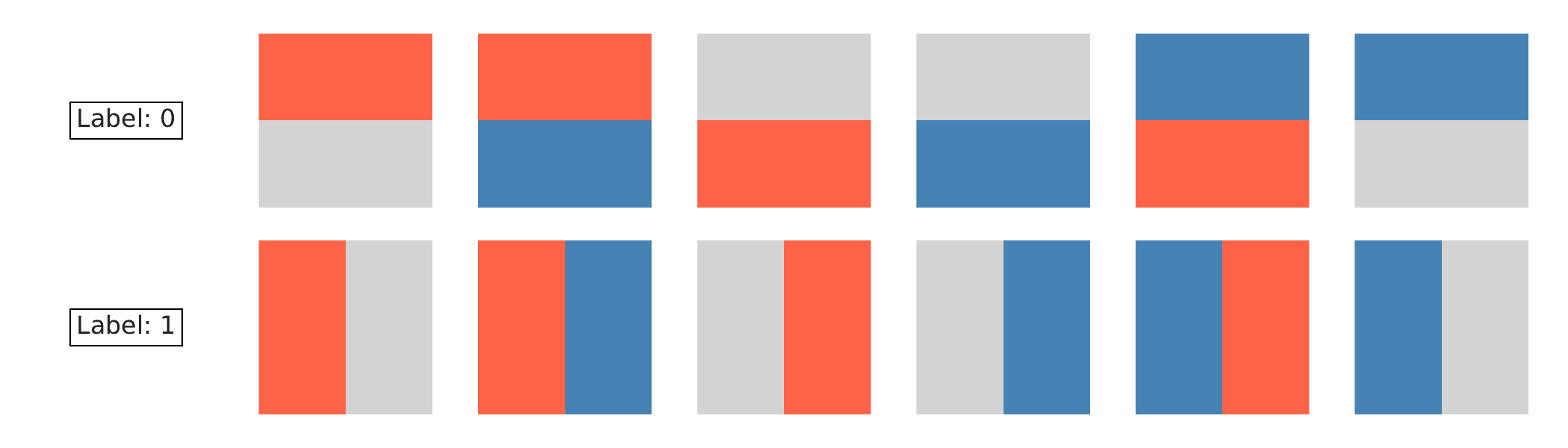}
    \caption{Dataset of the 12 binary images. The images are color-coded as follows: red represents a pixel value of -1, light gray represents a pixel value of 0, and blue represents a pixel value of 1. The labels are indicated on the left side of each row.}
    \label{fig:22dataset}
\end{figure}

\textbf{Dataset Generation.} A synthetic dataset was constructed comprising 12 binary images, each of size 2$\times$2 pixels, with corresponding binary labels (Fig. \ref{fig:22dataset}). The images were designed to represent simple patterns, such as adjacent pixel pairs with values of -1, 0, or 1. Labels were assigned based on the presence of specific patterns within the images, forming a binary classification task. 

This dataset has an important characteristic: it exhibits invariance under 180-degree rotations and flips. Specifically, if an image is rotated by 180 degrees or flipped, its classification label remains unchanged. However, if an image is rotated by 90 degrees, the classification label may change, indicating that the dataset does not exhibit 90-degree rotational invariance. This property is crucial for evaluating the effects of equivariance in neural networks.

\textbf{Model Architectures.} Five distinct network configurations were defined to isolate the effects of equivariance, dropout, and batch normalization. The detailed structures if these five networks are as following,

Baseline ConvNet: A basic convolutional neural network consisting of one convolutional layer, one fully connected layer, and an output layer.

Baseline + dropout: The baseline CNN augmented with a dropout layer between the convolutional and fully connected layers, with a dropout rate of 0.3.

Baseline + BatchNorm: The baseline CNN enhanced with a batch normalization layer following the convolutional layer.

Baseline + Equivariant: The baseline CNN modified to ensure convolutional filters are invariant under specific transformations, such as flips and 180-degree rotations.

Baseline + Wrong Equivariant: The baseline CNN adjusted to enforce invariance under 90-degree rotations, representing an incorrect equivariance constraint.

\textbf{Experimental Setup.} To comprehensively evaluate the loss landscapes, we generated all possible combinations of convolutional and linear layer weights, with each weight taking values of -1 or 1. 

In weight combination generation, all possible combinations of convolutional and linear layer weights were generated, with each weight taking values of -1 or 1.
For each weight combination, the five network configurations were used, and their loss values were computed using the cross-entropy loss function. 

\textbf{Loss Landscape Analysis.} To analyze the loss landscapes of different network configurations, we sorted the loss values for each weight combination to obtain a sequence of loss values from highest to lowest. 

\subsection{Measuring the Degree of Symmetry Breaking}

Measuring the degree of symmetry in a network is primarily achieved by comparing the correlations of the same network under different weight configurations (also called different replicas). We use different seeds to train the network models with identical hyper-parameters, and by measuring the symmetry (i.e., correlation) between replicas, we can equivalently measure the symmetry of the network structure itself. We applied this metric to the CIFAR-10 dataset using five distinct neural network architectures: SimpleCNN, DropoutCNN, BatchNormCNN, FlipEquivarianceCNN, and RotationEquivarianceCNN.

\textbf{Model Architectures.} Five distinct network configurations were defined to isolate the effects of equivariance, dropout, and batch normalization. The detailed structures of these five networks are as follows:

SimpleCNN: Consists of two convolutional layers followed by ReLU activations and max pooling, then a fully connected layer and an output layer.

DropoutCNN: Similar to SimpleCNN but includes a dropout layer with a dropout rate of 0.3 between the fully connected layer and the output layer.

BatchNormCNN: Similar to SimpleCNN but includes batch normalization layers after each convolutional layer.

FlipEquivarianceCNN: Similar to SimpleCNN but includes a data augmentation step in the forward pass that horizontally flips half of the images in the batch.

RotationEquivarianceCNN: Similar to SimpleCNN but includes a data augmentation step in the forward pass that rotates half of the images in the batch by 180 degrees.

\textbf{Training Different Networks.}

Data Preparation: The CIFAR-10 dataset was preprocessed by normalizing the images and splitting it into training and testing sets. Data augmentation techniques such as random horizontal flips and rotations were applied for FlipEquivarianceCNN and RotationEquivarianceCNN, respectively.

Model Training: The Adam optimizer with a learning rate of 0.001 and weight decay of $10^{-3}$ was used. Each model was trained for 30 epochs with a batch size of 64, using cross-entropy loss. Training losses and test accuracies were recorded for each epoch.

\subsection{Measuring the Degree of Symmetry Breaking}

Measuring the degree of symmetry in a network is primarily achieved by comparing the correlations of the same network under different weight configurations (also called different replicas). We use different seeds to train the network models with identical hyper-parameters, and by measuring the symmetry (i.e., correlation) between replicas, we can equivalently measure the symmetry of the network structure itself. We applied this metric to the CIFAR-10 dataset using five distinct neural network architectures: SimpleCNN, DropoutCNN, BatchNormCNN, FlipEquivarianceCNN, and RotationEquivarianceCNN.

\textbf{Model Architectures.} Five distinct network configurations were defined to isolate the effects of equivariance, dropout, and batch normalization. The detailed structures of these five networks are as follows:

SimpleCNN: Consists of two convolutional layers followed by ReLU activations and max pooling, then a fully connected layer and an output layer.

DropoutCNN: Similar to SimpleCNN but includes a dropout layer with a dropout rate of 0.3 between the fully connected layer and the output layer.

BatchNormCNN: Similar to SimpleCNN but includes batch normalization layers after each convolutional layer.

FlipEquivarianceCNN: Similar to SimpleCNN but includes a data augmentation step in the forward pass that horizontally flips half of the images in the batch.

RotationEquivarianceCNN: Similar to SimpleCNN but includes a data augmentation step in the forward pass that rotates half of the images in the batch by 180 degrees.

\textbf{Training Different Networks.}

Data Preparation: The CIFAR-10 dataset was preprocessed by normalizing the images and splitting it into training and testing sets. Data augmentation techniques such as random horizontal flips and rotations were applied for FlipEquivarianceCNN and RotationEquivarianceCNN, respectively.

Model Training: The Adam optimizer with a learning rate of 0.001 and weight decay of $10^{-3}$ was used. Each model was trained for 30 epochs with a batch size of 64, using cross-entropy loss. Training losses and test accuracies were recorded for each epoch.

\subsection{Measuring the Degree of Symmetry Breaking}

Measuring the degree of symmetry in a network is primarily achieved by comparing the correlations of the same network under different weight configurations (also called different replicas). We use different seeds to train the network models with identical hyper-parameters, and by measuring the symmetry (i.e., correlation) between replicas, we can equivalently measure the symmetry of the network structure itself. We applied this metric to the CIFAR-10 dataset using five distinct neural network architectures: SimpleCNN, DropoutCNN, BatchNormCNN, FlipEquivarianceCNN, and RotationEquivarianceCNN.

\textbf{Model Architectures.} Five distinct network configurations were defined to isolate the effects of equivariance, dropout, and batch normalization. The detailed structures of these five networks are as follows:

SimpleCNN: Consists of two convolutional layers followed by ReLU activations and max pooling, then a fully connected layer and an output layer.

DropoutCNN: Similar to SimpleCNN but includes a dropout layer with a dropout rate of 0.3 between the fully connected layer and the output layer.

BatchNormCNN: Similar to SimpleCNN but includes batch normalization layers after each convolutional layer.

FlipEquivarianceCNN: Similar to SimpleCNN but includes a data augmentation step in the forward pass that horizontally flips half of the images in the batch.

RotationEquivarianceCNN: Similar to SimpleCNN but includes a data augmentation step in the forward pass that rotates half of the images in the batch by 180 degrees.

\textbf{Training Different Networks.}

Data Preparation: The CIFAR-10 dataset was preprocessed by normalizing the images and splitting it into training and testing sets. Data augmentation techniques such as random horizontal flips and rotations were applied for FlipEquivarianceCNN and RotationEquivarianceCNN, respectively.

Model Training: The Adam optimizer with a learning rate of 0.001 and weight decay of $10^{-3}$ was used. Each model was trained for 30 epochs with a batch size of 64, using cross-entropy loss. Training losses and test accuracies were recorded for each epoch.

\textbf{Symmetry Metric Calculation.}

Training on 100 Images: A subset of 100 images from the CIFAR-10 dataset was selected. Each model was trained for 200 epochs under 200 different random seeds using the Adam optimizer, and the final weights were saved.

Dimensionality Reduction: The weights of each trained model were extracted and flattened. UMAP (Uniform Manifold Approximation and Projection) was then used to reduce the dimensionality of these weight vectors to 100 dimensions. UMAP is a non-linear dimensionality reduction technique that is particularly well-suited for preserving the global structure of high-dimensional data in a lower-dimensional space. This step helps in comparing the models by reducing the complexity of the weight space while retaining the essential features.

Pairwise Distance Calculation: The pairwise Wasserstein distance between the reduced feature representations of the trained models was computed to measure the minimum cost of transporting mass in one distribution to match another. This distance metric provides a meaningful way to compare the distributions of model weights.

Histogram and Mean Value Calculation: A histogram of the calculated Wasserstein distances was constructed, and Gaussian smoothing was applied to obtain a smooth curve. These smoothed histograms were normalized to form probability density functions. The mean value of the probability density was used as the metric. A higher mean value indicates a greater degree of symmetry breaking.

The Wasserstein distance provides a meaningful metric for comparing the distributions of model weights. By examining the mean value of the smoothed histogram, we can quantify the extent to which symmetry is broken in the trained models. A higher mean value suggests that the models have diverged more significantly in their weight distributions, indicating a higher degree of symmetry breaking.

\backmatter

\bmhead{Supplementary information}

Ablation study of input expansion in image classification.

\bmhead{Acknowledgements}
We are grateful to Dr. Jiarui Zhang in Northwest Institute of Nuclear Technology who provided suggestions on the clarity of the manuscript. This research has been funded
by the National Science Foundation of China under grant number 12105227, 12075098, 12435009, and 12405318.

\section*{Declarations}

\begin{itemize}
\item Competing interests: All authors declare no competing interests.
\item Code and data availability: The code and data used in the study can be found at https://doi.org/10.7910/DVN/QCYE4M.
\item Author contribution: 

\textbf{Jun-Jie Zhang:} Discovered the phenomenon of input expansion; proposed the symmetry breaking explanation; designed and conducted the experiments on dropout, batch normalization, and equivariance; developed the metric to measure symmetry.

\textbf{Nan Cheng and Xiu-Cheng Wang:} Discovered the phenomenon of input expansion; conducted all image classification tasks and the image coloring task.

\textbf{Fu-Peng Li:} Conducted the QCD experiments.

\textbf{Jian-Nan Chen:} Performed the example and analysis of the Ising model.

\textbf{Long-Gang Pang:} Suggested the test of the PINNacle; supervised the QCD experiments.

\textbf{Deyu Meng:} Suggested the development of the symmetry metric; supervised the entire work.

All authors contributed to the discussions, writing, revision, and figure preparation of the manuscript.

\end{itemize}


\bibliography{sn-bibliography}

\end{document}